%% file: main.tex
\def\year{2020}\relax
\documentclass[letterpaper]{article} 
\usepackage{aaai20}  
\usepackage{times}  
\usepackage{helvet} 
\usepackage{courier}  
\usepackage[hyphens]{url}  
\usepackage{graphicx} 
\usepackage{graphicx}  
\usepackage{multirow}

\usepackage{color}
\usepackage{soul}

\usepackage{amssymb}
\usepackage{amsmath}

\usepackage{verbatim}

\title{Evaluating the Effectiveness of Margin Parameter when Learning Knowledge Embedding Representation for Domain-specific Multi-relational Categorized Data}

\author{
  Matthew Wai Heng Chung\\
  HDR-UK Institute of Health Informatics\\
  University College London\\
  London, UK\\
  \texttt{wai.chung.18@ucl.ac.uk}
  \And
  Hegler Tissot\\
  HDR-UK Institute of Health Informatics\\
  University College London\\
  London, UK\\
  \texttt{h.tissot@ucl.ac.uk}
}

\newcommand{\hextrato}{{HEXTRATO}}

\begin{document}

\maketitle

%
%
\input{KER-LM-0-Abstract}
\input{KER-LM-1-Introduction}
\input{KER-LM-2-Background}

\input{KER-LM-3-Methodology}

\input{KER-LM-4-Results}

\input{KER-LM-6-Conclusions}

\bibliographystyle{aaai}
\bibliography{main}

\end{document}

%% file: KER-LM-0-Abstract.tex
\begin{abstract}
Learning knowledge representation is an increasingly important technology that supports a variety of machine learning related applications. However, the choice of hyperparameters is seldom justified and usually relies on exhaustive search.
Understanding the effect of hyperparameter combinations on embedding quality is crucial to avoid the inefficient process and enhance practicality of 
vector representation methods.
We evaluate the effects of distinct values for the margin parameter ($\gamma$) focused on translational embedding representation models for multi-relational categorized data. 
We assess the influence of $\gamma$ regarding the quality of embedding models by contrasting traditional link prediction task accuracy against a classification task.
The findings provide evidence that lower values of margin are not rigorous enough to help with the learning process, whereas larger values produce much noise pushing the entities beyond to the surface of the hyperspace, thus requiring constant  regularization.
%
Finally, the correlation between link prediction and classification accuracy shows traditional validation protocol for embedding models is a weak metric to represent the quality of embedding representation.
\end{abstract}
%

%% file: KER-LM-1-Introduction.tex
\section{Introduction}
\label{SECTION:Introduction}


Information extraction aims to recover facts from heterogeneous data, and attempts to capture that information using a multi-relational representation. The problem of representing multi-relational data has gained more attention in the last decade as more knowledge graphs become available and useful as supporting resources for a variety of machine learning related applications, such as information retrieval~\cite{Buttcher2010}, semantic parsing~\cite{Berant2013}, question-answering~\cite{Abujabal2017}, and recommender systems~\cite{Wang2019}.


A knowledge graph (KG) is a multi-relational dataset composed by entities (nodes) and relations (edges) that provide a structured representation of the knowledge about the world. Freebase~\cite{2008_Bollacker_Freebase}, Google Knowledge Graph~\cite{2014_Dong_GoogleKnowledgeVault}, Wordnet~\cite{1998_Fellbaum_Wordnet}, DBpedia~\cite{DBpedia2015}, and YAGO~\cite{2007_Suchanek_YAGO} are some well-known examples of KGs that provide reasoning ability and can be used for knowledge inference. 
However, the heterogeneous nature of the data sources, where facts are usually extracted from, makes the later typically inaccurate. Moreover, although containing a huge number of triplets, most of open-domain KGs are incomplete, covering only a small subset of the true 
knowledge domain they are supposed to represent.
Thus, learning the distributed representation of multi-relational data has been used as a tool to complete knowledge bases without requiring extra knowledge input. \emph{Knowledge base completion} or \emph{link prediction} (LP) refers to the problem of predicting new links (or new relationships) between entities by automatically recovering missing facts based on the observed ones.


Embedding methods are able to learn and operate on the latent feature representation of the constituents and on their semantic relatedness using an algorithm that optimizes a margin-based ($\gamma$) objective function over a training set. 
The problem of choosing the optimal combination of hyperparameters is a common practice usually tackled by performing an exhaustive ``grid-search" over several adjustable values before committing to a favorable training model. However, KGs can be very large and demand high computational costs to find the best configuration from all possible combinations of  hyperparameter options.


In this work, we focus on evaluating the effectiveness of choosing adequate values for the learning margin parameter $\gamma$ and how this choice is reflected on the accuracy of embedding models. There is no consensus on the optimal hyperparameters in previous work, and multiple approaches report best model accuracy with $\gamma$ ranging from 0.2 to 4.0. In addition, we contrast accuracy and quality of embedding representation by comparing the results between LP and classification tasks. 
Results provide strong evidence that lower values for the margin parameter are not necessarily rigorous enough to help with the learning process, whereas larger values produce much noise pushing the entities beyond to the surface of the hyperspace, causing entities to require constant regularization, which has a negative effect on the validation accuracy.
In addition, the correlation between LP and classification accuracy results shows traditional validation protocol for embedding models is a weak metric to represent the quality of embedding representation.
Finally, in order to advocate reproducibility and encourage the research community to test and compare novel embedding methods applied to multi-relational categorized data, we are making all the datasets used along our experiments available.

%% file: KER-LM-2-Background.tex
\section{Knowledge Embedding Representation}
\label{SECTION:Background}

A knowledge graph $\mathcal{G}$ is constructed with a set of facts $\mathcal{S}$ in the form of triples $(h,r,t)$. Each triple has a pair of head $h$ and tail $t$ entities $e \in \mathcal{E}$, connected by a relation $r \in \mathcal{R}$. Embedding methods aim to represent entities (\emph{h} and \emph{t}) and relations (\emph{r}) as vectors in a continuous $k$-dimensional vector space -- though most of enhanced models represent relations as \emph{k} x \emph{k} matrices.
%
%
%
Typically, embedding methods are able to learn and operate on the latent feature representation of the constituents, by defining a distinct relation-based scoring function $f_r(h,t)$ to measure the plausibility of the triplet $(h, r, t)$, where $f_r(h,t)$ implies a transformation on the pair of entities which characterizes the relation $r$. 

The final embedding representation is learned by optimizing a margin-based ($\gamma$) objective function (Equation \ref{equation:marginloss}) over a training set, while preserving the existing semantic relatedness among the triple constituents to enforce the embedding compatibility.
Non-existing negative triples $(h',r,t')$ are constructed for every observed triple in the training set by corrupting either the head or the tail entity. Both observed and corrupted triples are identically scored for comparison in the loss function, where $[x]_+ = max(0,x)$.


\begin{equation}\label{equation:marginloss}
\mathcal{L}=\displaystyle \sum_{\substack{(h,r,t)\in S\\
(h',r,t')\in S'}}{[\gamma - f_r(h,t) + f_r(h',t')]_+}
\end{equation}

TransE~\cite{2013_Bordes_TransE} is a baseline model in translational embedding representation learning. It presents a simple and scalable method to represent KGs in lower dimensional continuous vector space, though known for its flaws on representing one-to-many, many-to-one and many-to-many relations. 
%
In TransE, entities \emph{h}, \emph{t} and a relation \emph{r} are represented by translation vectors $\mathbf{h},\mathbf{t},\mathbf{r} \in \mathbb{R}^\emph{k}$, chosen so that every relation $r$ is regarded as a translation between $h$ and $t$ in the embedding space. The pair of embedded entities in a triple $(h, r, t)$ can be approximately connected by $\mathbf{r}$ with low error (Equation~\ref{equation:TransE}), and the plausibility score for an embedded triple is calculated by a function of distance measure between $\mathbf{h} + \mathbf{r}$ and $\mathbf{t}$ (Equation \ref{equation:TransE2}). 

\begin{equation}\label{equation:TransE}
\mathbf{h} + \mathbf{r} \approx \mathbf{t}
\end{equation}

\begin{equation}\label{equation:TransE2}
f_r(h,t) = \lVert h + r - t \rVert_{l_{1/2}}
\end{equation}

Later studies addressed several weaknesses of TransE and proposed extended models enhanced with additional features, including 
relation-specific projections to improve modelling of different data cardinality~\cite{2014_Wang_TransH,2015_Lin_TransR}, 
adapted scoring functions to allow more flexible translations~\cite{Fan2014TransM,Xiao2015TransA}, and 
Gaussian embeddings to model semantic uncertainty~\cite{He2015KG2E,Xiao2016TransG}.
More recent approaches attempted to improve KG completion performance by exploring several learning features, including 
tensor factorization~\cite{Nickel2011RESCAL},
compositional vector representation~\cite{Nickel2016HolE},
complex spaces~\cite{2016_Trouillon_Complex}, 
transitive relation embeddings~\cite{Zhou_Liu_Xu_Zhang_2019}, and 
neural neighborhood-aware  embeddings~\cite{Kong_Zhang_Mao_Deng_2019}.
%
%
However, there are multiple issues with these models that make them difficult for further development and their adoption in other domain-specific applications, including
(a) limited performance, (b) time-consuming validation processes, and (c) open-domain validation datasets that prevent embedding methods of taking advantage of more detailed and enriched metadata.
Indeed, 
\cite{Kadlec2017} cast doubt on
 the performance improvement claims
of more recent models whether being due to hyperparameter
tuning or different training objectives instead of  architectural changes, suggesting future research
to re-consider the way performance
of embedding models should be evaluated and reported.

\hextrato~\cite{hextrato2018} is a translational embedding approach that couples TransE with a set of ontology-based constraints to learn representations for multi-relational categorized data, originally designed to embed biomedical- and clinical-related datasets. In categorical datasets, each entity $e$ is associated with a category (or type) $c \in \mathcal{T}$. 
%
%
Designed to be used on embedding domain-specific data, \hextrato~improves the translational embedding by using typed entities that are projected onto type-based independent hyperspaces, achieving great performance on the LP task on an adapted (typed) version of Freebase, even in very low $k$-dimensional spaces ($k < 50$), without necessarily adding complex representation structures within the model training process. 


Given a training set $S$ of categorized triples \emph{($c_h$:$h$,$r$,$c_t$:$t$)}, \hextrato~learns embedding vectors for entities and relations, so that each categorized entity \emph{c:e} is represented by a embedding vector $e_c \in \mathbb{R}^k$, and each relation \emph{r} is represented by a embedding vector $r \in \mathbb{R}^k$. A score function $f_{r}$ (Equation \ref{HEXTRATO:score-function})  represents a L2-norm dissimilarity, such that the score $f_{r}(h_{c_h},t_{c_t})$ of a plausible triple \emph{($c_h$:$h$,$r$,$c_t$:$t$)} is smaller than the score $f_{r}(h'_{c_h},t'_{c_t})$ of an implausible triple \emph{($c_h$:$h'$,$r$,$c_t$:$t'$)}. Then, \hextrato~learns knowledge embedding representation by minimizing a margin-based ($\gamma$) loss function $\mathcal{L}$ (Equation \ref{HEXTRATO:loss-function-norm}) adapted from TransE, where $\gamma$ is the margin parameter, $\mathcal{S}$ is the set of correct triples, and $\mathcal{S}'$ is the set of incorrect triples \emph{($c_h$:$h'$,$r$,$c_t$:$t$) $\cup$ ($c_h$:$h$,$r$,$c_t$:$t'$)}. 

\begin{equation}\label{HEXTRATO:score-function}
f_{r}(h_{c_h},t_{c_t})=\lVert h_{c_h} + r - t_{c_t} \rVert_{l_{2}}
\end{equation}

\begin{equation}
\label{HEXTRATO:loss-function-norm}
\mathcal{L} = \sum_{\substack{(c_h:h,r,c_t:t) \in \mathcal{S}\\
(c_h:h',r,c_t:t') \in \mathcal{S}'}}{[\gamma + f_r(h_{c_h},t_{c_t}) - f_r(h'_{c_h},t'_{c_t})]_+}
\end{equation}

The combination of these 
constraint strategies diminishes the impact of two problems of model training discussed in literature: 
(a)  the ``zero loss" problem~\cite{Wang2018GANKRL,Shan2018NKRL} describes the observation that at later stages of training, corrupted triples tend to be sampled beyond the margin, which leads to a zero loss that is not useful for training 
-- this is because the corrupted head or tail entities are selected by random sampling and we expect false entities to gradually move away during training; %
and (b)  the ``false detection" problem~\cite{Shan2018NKRL}, which arises when poor-quality negative triples are constructed using entities that are often unrelated and have a different semantic type, which may give false confidence to the original triple and reduces representation accuracy.
In \hextrato, independent vector spaces for each type, coupled with restrictions on domain and range for each relation, lessen the probability of constructing a poor-quality negative triple.
The selection of corrupted triples is restricted by setting functional relations and disjoint sets instead of random sampling from the whole set of possible entities, so that training is more efficient and sped up, with reduced impact from uninformative training. 

HEXTRATO is proven to be faster than other embedding models due to the set of ontology-based constraints it uses, which optimizes the selection of negative samples during training. However it is still not efficient for a grid-search optimization. For real use-case scenarios, we aim to deploy embedding models in a more effective way, requiring the prototyping models with already known good (not necessarily the best) choice of hyperparameters, which has been the biggest motivation for this work.

%% file: KER-LM-3-Methodology.tex
\section{Methodology}
\label{SECTION:Methodology}

Domain-specific databases provide categorized data and metadata that can be used to enrich definitions of entities and relations within a knowledge base. Thus, each resulting triple is presented in the form \emph{($c_h$:h, r, $c_t$:t)}, where $c_h$ and $c_t$ represent the types of \emph{h} and \emph{t}. Besides providing  categorized entities, relations are also restricted by domain and range. 
In the following example, the relation \emph{hasGender} is constrained by the domain \emph{patient} and the range \emph{gender} -- in addition, using independent vector spaces to project each entity type leads to a substantial processing time improvement along the validation process.

\ 

{\centering
\texttt{(patient:P01, hasGender, gender:male)}
}

\

We aim to evaluate the effectiveness of the 
margin parameter $\gamma$ when learning embedding representation for domain-specific multi-relational categorized data along distinct scenarios regarding dimensionality ($k$) and dataset sizes and shapes (relation cardinality). 
Thus, we used the primary ontological constraint proposed by HEXTRATO (typed entities) in order to train a set of embedding models that allowed us to contrast the target hyperparameter performance - this approach is reported in the original paper as ``H1".

We believe dataset shapes can considerably affect the performance of embedding methods, as more complex embedding representation models tend to adapt the way relations are taken into account in order to improve accuracy within the LP task. However, the entity representation is just marginally affected. Thus, this work focuses on embedding quality assessment instead of trying to improve LP performance.

\subsection{Datasets}

HEXTRATO was originally evaluated using the following datasets:
%
(a) two real clinical-related datasets extracted from \emph{InfoSaude}~\cite{Tissot2018_MND} -- an Electronic Health Record (EHR) system;
(b) Mushroom, a publicly available dataset deposited on the UCI Machine Learning Repository
that describes hypothetical samples corresponding to distinct species of mushrooms; and
(c) an adapted version of FB15K dataset (FB15K-Typed), which has been simplified to a set of distinct 55 types in this work (FB55T).
In addition, we also included a new dataset (BPA) that is presented as set of restrictions on how medical procedures are constrained by distinct diagnosis. An overview of each dataset\footnote{\url{https://github.com/hextrato/KER/tree/master/datasets}} is presented below and statistics are depicted in Table~\ref{tab:datasets-stats}.

{\bf EHR-Demographics}
comprises a set of 2,185 randomly selected patients from the \emph{InfoSaude} system who had at least one admission between 2014 and 2016. Each patient is described by a set of basic demographic information, including gender, age (range in years) in the admission, marital status (unknown for about 15\% of the patients), education level, and two flags indicating whether the patient is known to be either a smoker or pregnant, and the social groups assigned according to a diverse set of rules mainly based on demographic and historical clinical conditions. Demographic features are represented by \emph{many-to-one} relations, whereas association of each patient to social groups is given by a \emph{many-to-many} relation.

{\bf EHR-Pregnancy}
is a dataset used to identify correlations between pre- and post-clinical conditions on pregnant patients with abnormal pregnancy termination, comprised by a set of 2,879 randomly selected pregnant female patients from the \emph{InfoSaude} system in which pregnancy was inadvertently and abnormally interrupted before the expected date of birth; each patient is described by age (range in years),  known date of last menstrual period (LMP), whether the patient had an abortion (regardless of reason), and a list of \mbox{ICD-10} (the 10th revision of the International Classification of Diseases) codes~\cite{WHOICD10} registered either before or after the LMP date. This is a dataset mostly comprising \emph{many-to-many} relations that connects patients with corresponding diagnoses.

{\bf Mushroom}
uses a set of features to describe 8,124 hypothetical species of mushrooms, including shape, surface, color, bruises, odor, gill, stalk, veil, ring, spore, population, and habitat - intended to identify whether each species is  edible or poisonous given its featured characteristics. All features within this dataset are given by \emph{many-to-one} relations.

{\bf BPA} (Ambulatory Production Bulletin)\footnote{\url{http://datasus.saude.gov.br/sistemas-e-aplicativos/ambulatoriais/sia}}
is an outpatient care dataset that allows the service provider to be linked to the Public Health Ministry in Brazil to record the care performed at the health facility on an outpatient basis; in order to optimize the data remittance process, there are several rules for the correct completion of submitted data that must be followed strictly, including the restrictions between medical procedures and their constrained diagnoses from \mbox{ICD-10}. This dataset associates medical procedures with a multilevel hierarchical set of \emph{many-to-one} relations, coupled with \emph{many-to-many} relations that impose multiple restrictions on each procedure, mainly regarding to possible related diagnoses.

{\bf FB55T}
is adapted from the Freebase-Typed dataset used along HEXTRATO experiments, from which each entity was categorized based on the metadata description of their corresponding relations; in FB55T, we reduced the high cardinality of 1,248 hierarchical types to a set of 55 aggregated types, each type mostly acting as an entity domain context, though over 50\% of the triples have distinct aggregated types between head and tail.

\begin{table*}
\caption{Statistics of domain-specific benchmark datasets, given by the number of entities, relations, types, and triples in each dataset split -- training (LRN), validation (VLD), tuning (TUN) and test (TST) sets.}
\label{tab:datasets-stats}
\center
\begin{tabular} { | p{2.5cm} | p{2.5cm} | p{2.5cm} | p{2.5cm} | p{2.5cm} | p{3.5cm} | }
\hline											
		&	\multicolumn{5}{c|}{{\bf Datasets }}	\\
\cline{2-6} 
		& \multicolumn{1}{c|}{{\bf EHR}}	& \multicolumn{1}{c|}{{\bf EHR}}  & \multicolumn{1}{c|}{{\bf UCI}} & \multicolumn{1}{c|}{~} & \multicolumn{1}{c|}{{\bf Freebase}}  \\
{\bf \# (number of)}		&	\multicolumn{1}{c|}{{\bf Demographics}}	&	\multicolumn{1}{c|}{{\bf Pregnancy}}   & \multicolumn{1}{c|}{{\bf Mushroom}}  & \multicolumn{1}{c|}{{\bf BPA}} & \multicolumn{1}{c|}{{\bf FB55T}} \\
\hline											

Entities 		& \multicolumn{1}{r|}{2,237}	& \multicolumn{1}{r|}{3,088	}	& \multicolumn{1}{r|}{8,487} 	& \multicolumn{1}{r|}{22,874}  	& \multicolumn{1}{r|}{14,951 }	\\
\hline											
Relations		& \multicolumn{1}{r|}{6}		& \multicolumn{1}{r|}{5}		& \multicolumn{1}{r|}{23} 		& \multicolumn{1}{r|}{23  }		& \multicolumn{1}{r|}{1,345}  	\\
\hline											
Types			& \multicolumn{1}{r|}{7}		& \multicolumn{1}{r|}{4}		& \multicolumn{1}{r|}{15 } 		& \multicolumn{1}{r|}{14 }		& \multicolumn{1}{r|}{55 	}	\\
\hline											
Triples (total) & \multicolumn{1}{r|}{15,345}	& \multicolumn{1}{r|}{20,768}	& \multicolumn{1}{r|}{191,088}	& \multicolumn{1}{r|}{186,177}	& \multicolumn{1}{r|}{592,213}	\\
\ \ \ \ \ LRN	& \multicolumn{1}{r|}{13,875}	& \multicolumn{1}{r|}{14,588}	& \multicolumn{1}{r|}{153,057}	& \multicolumn{1}{r|}{177,727}	& \multicolumn{1}{r|}{483,142}	\\
\ \ \ \ \ VLD	& \multicolumn{1}{r|}{463}		& \multicolumn{1}{r|}{1,997	}	& \multicolumn{1}{r|}{9,525 }	& \multicolumn{1}{r|}{2,889 }	& \multicolumn{1}{r|}{50,000 }	\\
\ \ \ \ \ TUN	& \multicolumn{1}{r|}{475}		& \multicolumn{1}{r|}{2,093}	& \multicolumn{1}{r|}{9,564 }	& \multicolumn{1}{r|}{2,729 }	& \multicolumn{1}{r|}{n/a 	}	\\
\ \ \ \ \ TST	& \multicolumn{1}{r|}{532}		& \multicolumn{1}{r|}{2,090}	& \multicolumn{1}{r|}{18,942 }	& \multicolumn{1}{r|}{2,832} 	& \multicolumn{1}{r|}{59,071 }	\\
\hline											
\end{tabular}
\end{table*}

\subsection{Evaluation Protocol} 


For each dataset, we look for the best embedding model corresponding to each possible pair of hyperparameters $(\gamma,k)$, aiming to analyze the impact of distinct values for the margin parameter $\gamma \in \{0.25, 0.5, 1.0, 1.5, 2.0, 3.0, 4.0\}$ in distinct $k$-dimensional spaces $k \in \{32,64,128\}$, regarding the resulting embedding representation accuracy. Our evaluation protocol comprises two main steps: (a) we primarily assess the accuracy of the knowledge embedding representation model using LP as a reference for accuracy performance in order to compare the effects of choosing distinct values for the margin parameter $\gamma$; (b) then, the quality of the resulting embedding representation is evaluated by a classification task, designed as multi-categorical classification problem for each one of the evaluation datasets. Finally, the correlation between the accuracy in the LP and classification tasks is analyzed in order the evaluate the effectiveness of $\gamma$ regarding the quality of each resulting embedding model. 

The following training protocol was proposed by HEXTRATO, in which multiple learning processes are used to train independent replicas initialized with random vector representations in order to find the best configuration set:

\begin{itemize}
\item Given a training set $S$ of triplets \emph{($c_h$:h, r, $c_t$:t)} we trained embedding models to learn vector representation for each categorized entities and relation. Entity and relation embedding vectors are initialized with the random uniform normalized initialization~\cite{AISTATS2010_GlorotB10}. The set of golden triples is then randomly traversed multiple times along the training process,
such that each training step produces a corrupted triple for each correct triple.

\item The dissimilarity measure was set to the L2-norm distance based on $\gamma$, and each optimal model setup is determined by early stopping accordingly to the MRR accuracy on the validation set. Ten distinct replicas of each model are independently trained for each set of targeted hyperparameters. After traversing all the training triplets at most 1,000 epochs with a learning rate $\lambda = 0.01$, the best model is chosen by comparing the scores against a tuning set -- the tuning set is used to choose the best replica. 

\item Final resulting scores are then calculated over the test set. FB55T has no tuning set, so that we were only able run one single replica for each target hyperparameter pair setup.

\item Incorrect triples $(c_h:h', r, c_t:t) \cup (c_h:h, r, c_t:t')$ are generated by randomly corrupting either \emph{h} or \emph{t} in a correct triple $(c_h:h,r,c_t:t) \in \mathcal{S}$ using a uniform probability for entity replacement, in which the entity replacement is randomly chosen from the set of entities belonging to the corresponding type of each original relation domain and range. We used Stochastic Gradient Descent (SGD)~\cite{1951_SGD_Robbins} to minimize the margin-based loss functions $\mathcal{L}$.

\item Within the resulting model, each entity \emph{c:e} is represented by a embedding vector $e_c \in \mathbb{R}^k$, and each relation \emph{r} is represented by a embedding vector $r \in \mathbb{R}^k$. Similarly to TransE, there is a score function $f_{r}$ regarding each relation $r$ (Equation \ref{HEXTRATO:score-function}) that represents a dissimilarity metric, where the score $f_{r}(h_{c_h}, t_{c_t})$ of a plausible triple \emph{($c_h$:h, r, $c_t$:t)} is smaller than the score $f_{r}(h'_{c_h}, t'_{c_t})$ of an implausible triple \emph{($c_h$:h', r, $c_t$:t')}. 
\end{itemize}


For evaluation purposes, we designed LP as a question answering task which aims at completing a categorized triple $(c_h:h, r, c_t:t)$ with \emph{h} or \emph{t} missing, though the type or category $c_h$ or $c_t$ of the missing entity is known. The plausibility of a set of candidate entities in descending order of similarity scores records the rank of the correct missing entity according to the corresponding entity type constraint. LP results are reported by Mean Reciprocal Rank (MRR), whereas achieving higher MRR is taken as good link predictor scores. LP results are presented in Figure~\ref{fig:Heatmaps}~(a,b).


Finally, each embedding model was submitted to a classification task. We selected one of the relations in each knowledge base as the target relation for each classifier -- the choice was made by looking at the relation with less missing values regarding the set of head entities. Each classifier was designed with: (a) input layer size equals $k$, (b) hidden intermediate layer size equals $2 \times k$, and (c) output layer size equals the number of possible classes. We used logistic activation functions in all hidden and output cells. The set of relations corresponding to the target relation was split into training (80\%), validation (10\%) and test (10\%) sets, and each classifier was submitted up to 1,000 training epochs early stopping accordingly to the classification accuracy on the validation set. After training at most 1,000 epochs with a learning rate $\lambda = 0.01$, the final classification scores are then calculated over the test set. Each result embedding model as submitted to the same classification protocol in order to evaluate effectiveness of resulting vector representation as a measure of quality. Classification results are presented in Figure~\ref{fig:Heatmaps}~(c). 

We believe the practical usage of this methodology is as important which was not thoroughly explored in previous publications. Instead of introducing new parameters, we aim to improve our understanding of the parameters that determine the functionality and accuracy of embedding models. Moreover, our evaluation by classification tasks propose a more realistic approach to training models for categorized datasets and challenges the current notion of a better LP score implies a better representation model.

%% file: KER-LM-4-Results.tex
\section{Results}
\label{SECTION:Results}

\begin{figure*}
     \centering
        (a) MRR \\
        \includegraphics[width=\textwidth]{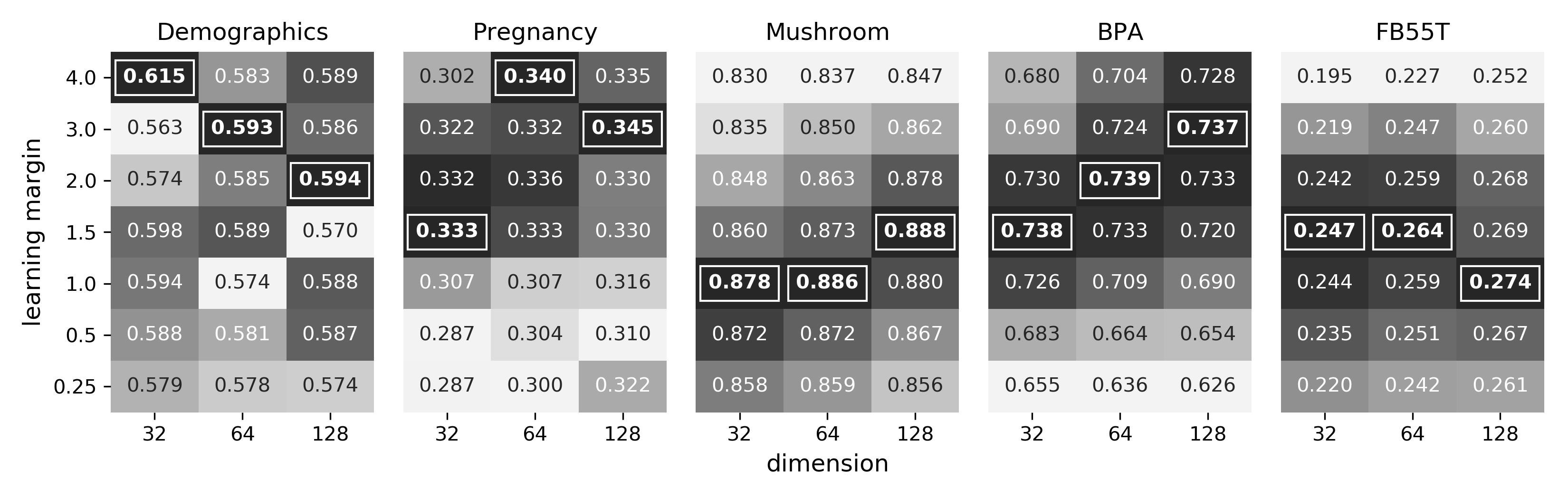}
        \\ (b) MRR$_r$ \\
        \includegraphics[width=\textwidth]{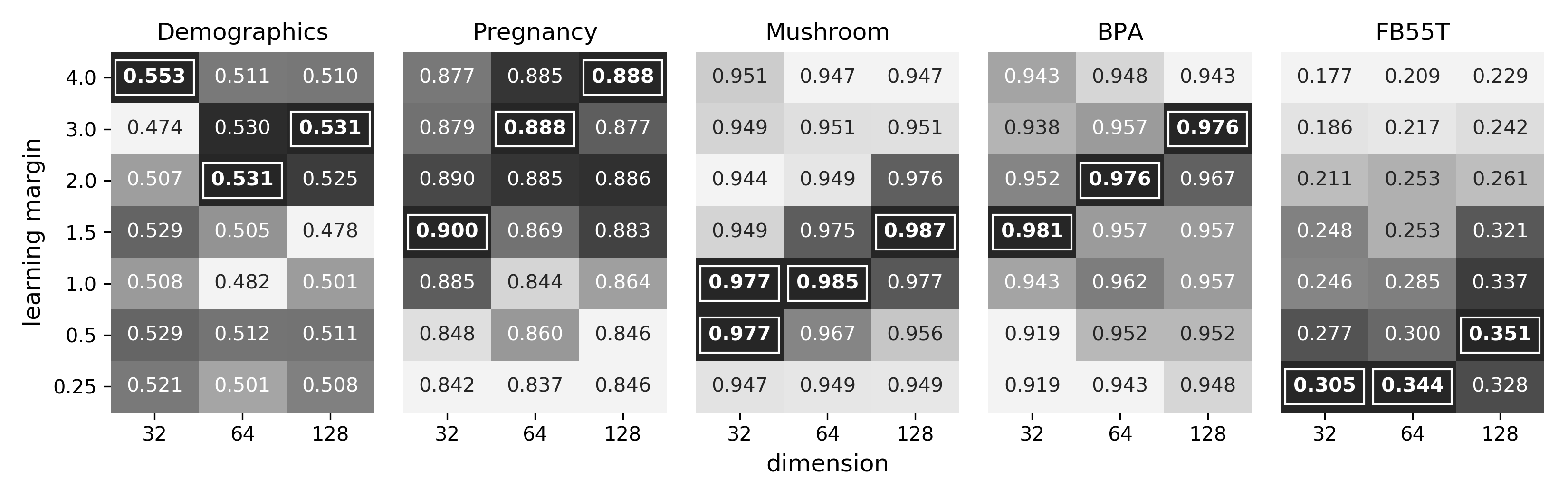}
        \\ (c) Classifiers \\
        \includegraphics[width=\textwidth]{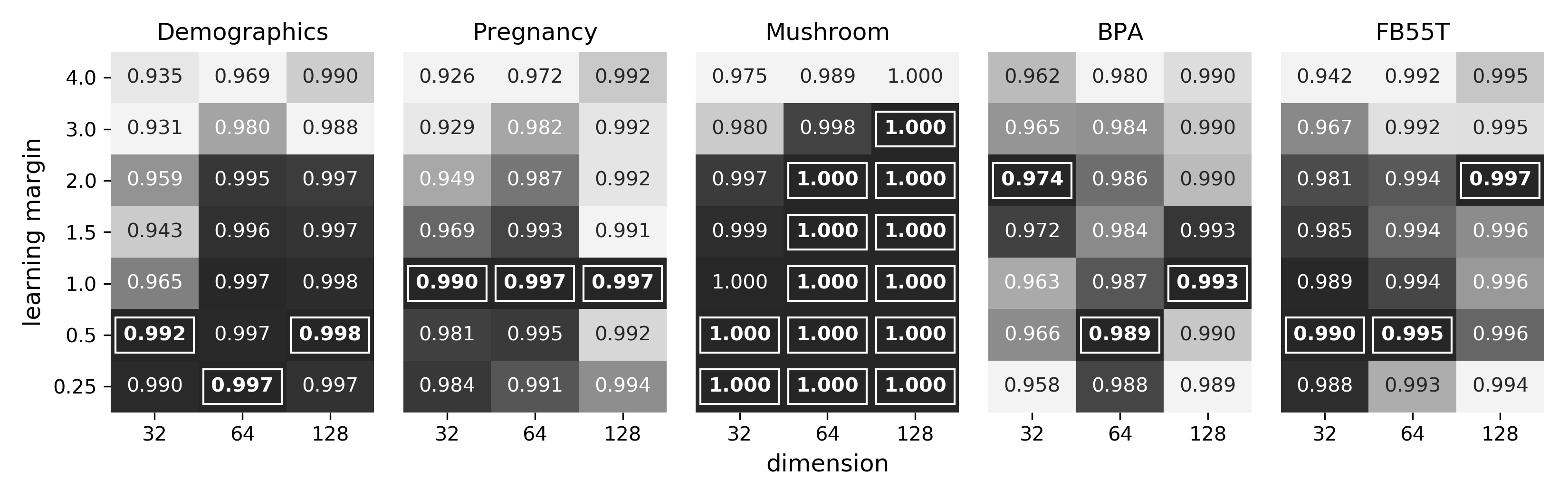}
        \caption{LP and classification metrics by learning margin and dimensionality ($k$) of each model. \textbf{(a)} overall MRR; \textbf{(b)} MRR for the target relation (MRR$_r$) used along the classification task; \textbf{(c)} classifier accuracy. The boxes are colored according to the scale of each column, ranging from the worst values in the lightest grey to the best values in the darkest grey. 
        The metric score of each pair $(\gamma,k)$ is annotated in the cell, and the best score is highlighted in a box. 
        }
        \label{fig:Heatmaps}
\end{figure*}

The primary objective of this work is to assess the choice of learning margin on the accuracy of learned embedding models, particularly in the context of multi-relational categorized data. 
Data quality, ambiguity of category definitions and missing data erode the knowledge represented and undermine the embedding model. However, as long as we are making comparisons using the same dataset their effects would not confound our findings.

Figure \ref{fig:Heatmaps} presents MRR, MRR for the classifier target relation (MRR$_r$), and classification accuracy for
all pairs $(\gamma,k)$.
%
%
In general, a margin between 1.0 and 2.0 is preferred in terms of LP across Mushroom, BPA, and FB55T datasets, whereas for EHR Demographics and Pregnancy larger values between 2.0 and 4.0 tend to provide better accuracy in LP. 
Larger margins tend to work better in higher $k$-dimensional spaces when comparing results within the same dataset. A plausible reason is that higher dimensions give more room to enforce the margin onto the entities.
However, in lower $k$-dimensional spaces, larger values of $\gamma$  may create more noise, pushing entities towards the hyperspace surface. Conversely, setting a very small margin may not be adequate to improve the representation and is therefore less likely to reach the best LP accuracy.
%
Still regarding LP, MRR scores on different dimensions are similar, so that higher dimensions do not necessarily outperform lower ones, which led us to conclude higher $k$-dimensional spaces may be not required to model larger datasets (e.g. FB55T), but  are more related to improving performance regarding the semantic complexity of the knowledge graph (e.g. BPA).


Along the classification task, all the classifiers achieved high accuracy (Figure \ref{fig:Heatmaps}c), regardless of the range of MRR.
Although linear embedding models, like TransE and HEXTRATO, cannot directly capture the complexity of non-separable problems, their resulting embedding representation is somehow able to capture the semantic representativeness of entities and their relations, 
directly reflecting on high accuracy when the vector representation is used as input for more complex machine learning classification tasks.
Interestingly, in mostly all the scenarios, an embedding model trained with a smaller margin tends to allow classifiers to achieve higher accuracy than those using embedding models trained with a larger margin.


Finally, for each $k$, we calculated Pearson's and Spearman's rank-order correlation coefficients to analyze the relationship between LP and classification metrics (Table \ref{table:correlation-coefficients}).
In most cases, there are some correlation, though not always strong.
When correlation exists, the direction is usually positive for MRR although the opposite is observed in \textit{Demographics}, \textit{Pregnancy} and  \textit{BPA} with $k=64$. 
LP and classification metrics do not fully agree on the overall quality of embedding representation, as their correlation is inconsistent and unreliable.
The correlation between MRR$_r$ and classifier accuracy follows the same direction as MRR but the strength often varies. Despite focusing on the same relation, in most cases MRR$_r$ does not have strong correlation with classifier accuracy. This implies that the two evaluation tasks are not directly related and it is inaccurate to infer one from another.

\begin{table*}[t]
\centering
\caption{Pearson and Spearman correlation coefficients between LP metrics and corresponding model classifier accuracy for distinct model dimensionality $k \in \{32,64,128\}$ -- MRR$_r$ stands for the equivalent MRR of the classifier target relation.}
\label{table:correlation-coefficients}
\begin{tabular}{|c|r|r|r|r|r|r|}
\hline
\multicolumn{1}{|c|}{\multirow{2}{*}{\bf{Datasets}}} & \multicolumn{2}{c|}{{\bf k = 32}} & \multicolumn{2}{c|}{{\bf k = 64}} & \multicolumn{2}{c|}{{\bf k = 128}} \\ \cline{2-7} 
\multicolumn{1}{|l|}{} &  \multicolumn{1}{c|}{{\bf MRR}} & \multicolumn{1}{c|}{{\bf MRR$_r$}} &  \multicolumn{1}{c|}{{\bf MRR}} & \multicolumn{1}{c|}{{\bf MRR$_r$}} &  \multicolumn{1}{c|}{{\bf MRR}} & \multicolumn{1}{c|}{{\bf MRR$_r$}} \\ \hline
Demographics 	&  -0.137 & 0.156 &  -0.385 & -0.344 &  -0.264 & -0.507 \\ \hline
Pregnancy 		&  -0.407 & -0.347 & -0.695 & -0.708 &  -0.500 & -0.420 \\ \hline
Mushroom 		&  0.867 & 0.346 &  0.764 & 0.444 &  0.654 & 0.442 \\ \hline
BPA 			&  0.818 & 0.702 & -0.471 & 0.094 &  0.211 & 0.160 \\ \hline
FB55T 			&  0.843 & 0.804 &  0.757 & 0.508 &  0.563 & 0.146 \\ \hline
\multicolumn{7}{c}{(a) Pearson correlation coefficients} \\ 
\multicolumn{7}{c}{~} \\
\hline 
\multicolumn{1}{|c|}{\multirow{2}{*}{\bf{Datasets}}} & \multicolumn{2}{c|}{{\bf k = 32}} & \multicolumn{2}{c|}{{\bf k = 64}} & \multicolumn{2}{c|}{{\bf k = 128}} \\ \cline{2-7} 
\multicolumn{1}{|l|}{} &  \multicolumn{1}{c|}{{\bf MRR}} & \multicolumn{1}{c|}{{\bf MRR$_r$}} &  \multicolumn{1}{c|}{{\bf MRR}} & \multicolumn{1}{c|}{{\bf MRR$_r$}} &  \multicolumn{1}{c|}{{\bf MRR}} & \multicolumn{1}{c|}{{\bf MRR$_r$}} \\ \hline
Demographics 	& 0.000 & 0.107 &  -0.714& -0.643 &-0.143 & -0.464 \\ \hline
Pregnancy 		& -0.286 & -0.143&-0.607 & -0.721 &-0.487 & -0.400 \\ \hline
Mushroom 		& 0.793 & 0.174 &  0.802 & 0.539 & 0.612 & 0.612 \\ \hline
BPA 			& 0.821 & 0.564 & -0.464 & -0.090 & 0.286 & 0.631 \\ \hline
FB55T 			& 0.536 & 0.821 &  0.607 & 0.607 & 0.536 & 0.250 \\ \hline
\multicolumn{7}{c}{(b) Spearman correlation coefficients} 
\end{tabular}
\end{table*}

In general, we observed that a learning margin between 0.5 and 1.5 results in consistently better quality of embedding representation. 
Larger values push the entities beyond the surface of the hyperspace which could produce much more noise than setting the regularization to $|x| = 1$, as opposed to $|x| \le 1$ as in TransE~\cite{2013_Bordes_TransE} and some other models~\cite{Fan2014TransM,2011_Bordes_SE}. In contrast, resulting accuracy from lower margin values ($\gamma < 0.5$) is frequently inferior possibly because the margin is not rigorous enough to help with the learning process.

Based on the classification results, we have strong evidence that the LP metrics do not directly reflect the effectiveness of using the resulting vector representation in specific classification tasks. While there may be some overall correlation between the metrics, the inconsistency makes it unreliable to base our judgment entirely on MRR in these situations.
While the classifiers are focused on a very specific group of triples from each KG, these experiments allowed us to reflect on the extent of extrapolating LP performance to possible subsequent tasks. Especially in specific domains with abundant categorized data such as in Electronic Health Records, embedding representation may have more applications in classifiers for clinical decision support rather than knowledge graph completion, whereas the latter has been a major research motivation in the open domain.

%% file: KER-LM-6-Conclusions.tex
\section{Conclusions}
\label{SECTION:Conclusions}

This study focuses on the learning margin parameter as we observed several patterns of margin preference according to the dataset characteristics, model dimensionality, and  evaluation task. Our findings provide preliminary evidence to understand the choice of hyperparameters in the context of learning representation for  multi-relational categorized data, an area which lacks formal justification for hyperparameter optimization. 

We trained and evaluated an embedding model with a range of learning margins while holding other hyperparameters constant on multiple categorized datasets with various sizes and shapes. Additionally, we analyzed multiple resulting models with different embedding vector dimensions to investigate any relationship between $\gamma$ and $k$ and their possible interactions. The evaluation is primarily focused on the standard LP task. However, subject to the intended usage of the embeddings such as for building a classifier, we question whether LP can directly reflect their effectiveness. Classifiers that predict a targeted relation in each knowledge graph are trained as a secondary evaluation task to assess the appropriateness of LP metrics.

Based on experimental results, we provide evidence that lower values for the margin parameter are not necessarily rigorous enough, whereas larger values produce much noise pushing the entities beyond to the surface of the hyperspace, leading to frequent regularization. More importantly, usual LP metrics do not necessarily represent the quality of resulting vector representation, as the correlation between LP and classification accuracy metrics tends to be weak.

Some of the ways in which this work can be extended include:

(a) Evaluating other combination of hyperparameters with respect to the data and task at hand, as such findings could be useful for working around the reliance on exhaustive grid search which is ineffective and is a huge barrier against widespread application of embedding representation for categorized multi-relational data.

(b) Gathering further evidence about the effectiveness of the learning margin regarding other ontological constraints as well as expanding the scope to other hyperparameter sets.

(c) Towards finding alternative ways of accurately evaluating an embedding representation model that corresponds to its intended use, we plan to test whether and in what extent clustering is able to replace traditional LP metrics within the embedding training and evaluation protocols.

(d) About the generalizability of our results, we may be still able to fit in a few more evaluation tasks.

Although robust conclusions cannot be drawn from these experiments performed with a single model which is an extension of TransE,  we try to provide evidences that embedding models can be differently affected with the choice of hyperparameters. A more meaningful experimental setup must be considered in order to verify whether reported correlation results are a dataset shape-dependent phenomenon and lead to a more general conclusion.
Now that we have a starting point in hyperparameter association and the implication of different evaluation tasks, we can scale up our experiments to include more datasets with variable sizes and shapes as well as using multiple evaluation tasks; however, we believe our current results adequately represent the trend in learning margin versus dimension and the difference between link prediction and classification metrics.

Finally, although HEXTRATO is a promising approach to learn knowledge embedding representation for multi-relational categorized data, there is still room for improvement. We plan to explore various additional modelling strategies and hyperparameter options including further ontology-based constraints, the use of projection matrices and distinct regularization constraints. 


